\documentclass[conference]{IEEEtran}
\IEEEoverridecommandlockouts
\usepackage[
    backend=biber,
    style=numeric,
    maxcitenames=1,
    maxbibnames=1,
    backref=True,
    doi=false,isbn=false,url=false,eprint=false
]{biblatex}
\usepackage{amsmath,amssymb,amsfonts}
\usepackage{algorithmic}
\usepackage{graphicx}
\usepackage{textcomp}
\usepackage{float}
\usepackage{caption}
\usepackage{subcaption}
\usepackage{multirow}
\usepackage{url}
\usepackage[table,xcdraw]{xcolor}
\usepackage{booktabs}
\usepackage{soul,xcolor}
\usepackage[normalem]{ulem}
\usepackage{xcolor}
\usepackage{todonotes}
\usepackage{afterpage}
\usepackage[final]{changes}
\usepackage{xcolor}
\usepackage{todonotes}
\usepackage{hyperref}

\definechangesauthor[color=brown, name={Milad Leyli-abadi}]{ML}
\definechangesauthor[color=purple, name={jean-patrick}]{JP}
\definechangesauthor[color=orange, name={Ali tahmasebi}]{AT}
\definechangesauthor[color=green, name={Sonja Van Ockenburg}]{SO}

\def\BibTeX{{\rm B\kern-.05em{\sc i\kern-.025em b}\kern-.08em
    T\kern-.1667em\lower.7ex\hbox{E}\kern-.125emX}}

\newcommand\redsout{\bgroup\markoverwith{\textcolor{red}{\rule[0.5ex]{2pt}{0.4pt}}}\ULon}

\begin{document}
\setstcolor{red}
\title{Statistical and Predictive Analysis to Identify Risk Factors and Effects of Post COVID-19 Syndrome}



\author{\IEEEauthorblockN{Milad Leyli-abadi, Jean-Patrick Brunet, Axel Tahmasebimoradi}
\IEEEauthorblockA{
IRT SystemX, Palaiseau, France\\
milad.leyli-abadi@irt-systemx.fr
}}



\maketitle

\begin{abstract}

Based on recent studies, some COVID-19 symptoms can persist for months after infection, leading to what is termed long COVID. Factors such as vaccination timing, patient characteristics, and symptoms during the acute phase of infection may contribute to the prolonged effects and intensity of long COVID. Each patient, based on their unique combination of factors, develops a specific risk or intensity of long COVID. In this work, we aim to achieve two objectives: (1) conduct a statistical analysis to identify relationships between various factors and long COVID, and (2) perform predictive analysis of long COVID intensity using these factors. We benchmark and interpret various data-driven approaches, including linear models, random forests, gradient boosting, and neural networks, using data from the Lifelines COVID-19 cohort. Our results show that Neural Networks (NN) achieve the best performance in terms of MAPE, with predictions averaging 19\% error. Additionally, interpretability analysis reveals key factors such as loss of smell, headache, muscle pain, and vaccination timing as significant predictors, while chronic disease and gender are critical risk factors. These insights provide valuable guidance for understanding long COVID and developing targeted interventions.

\end{abstract}


\section{Introduction}
In may 2023, after 3 years of global pandemic, the WHO declared the end of the global Public Health Emergency for COVID-19. While this indicates an improvement, notably with the general access to vaccines, it does not mean the end of the presence and effects of COVID-19 \cite{the_lancet_covid-19_2023}. One such lasting effects being for a large proportion of patients to continue presenting with physical and cognitive symptoms after recovery from acute COVID-19 \cite{srikanth_identification_2023}.
This condition being described as long-haul COVID-19, Long Covid (LC), or post-acute sequela of COVID-19 (PASC). With an estimated 43.9\% of the world population having been contaminated between the start of the pandemic and November 2021 \cite{barber_estimating_2022} this could have a lasting impact both at the individual and societal level.


Many research works aimed to use ML models to analyze and investigate COVID-19 infection. In the survey done by Vaishya et al \cite{vaishya2020artificial} in 2020, ML models were used for 7 critical areas to monitor and control the COVID-19 pandemic. The main applications of AI in this medical domain are: (1) early detection and diagnosis, (2) treatment monitoring, (3) identifying COVID-19 clusters, (4) mortality prediction and projection of spread and infection, (5) vaccine development, (6) reducing the workload of healthcare system, and (7) prevention. In the following, we present the related works that encompass one or more of these application domains.


Adamidi et al. \cite{adamidi2021artificial} surveyed Machine Learning (ML) models for COVID-19 screening, diagnosis, and prognosis which used various source of knowledge such as medical imaging, clinical parameters, lab results, and demographic features. They have conducted a risk of bias analysis to determine the applicability of the studies in clinical settings and to help healthcare professionals, those creating guidelines, and decision-makers.


Dairi et al. \cite{dairi2021comparative} studied machine learning methods for short-term COVID-19 transmission forecasting. They compared multiple deep learning models such as hybrid Long Short-Term Memory Convolutional Neural Networks, hybrid gated recurrent unit-convolutional neural networks 
and their non-hybrid counterparts with machine learning baseline methods like logistic regression and support vector regression. 
The performance was tested using confirmed and recovered COVID-19 cases time-series data from seven impacted countries and the results reveal that hybrid deep learning models can efficiently forecast COVID-19 cases with a better performance than baseline models.

In 2022, Orooji et al. \cite{orooji2022comparing} compared several training algorithms for a Multi-layer Perceptron (MLP) to predict the length of stay of COVID-19 patients. From 53 total features, 20 were selected based on a correlation analysis. Interestingly, features like nausea, vomiting, headache, muscle pain, chills, fever, loss of taste, loss of smell, and sore throat were excluded. The dataset included 1,225 patients, with 10\% used for testing and 90\% for training the neural networks. They concluded that the MLP of 10 hidden neurons with Bayesian Regularization training algorithm yielded the best results.

Kumar et al. \cite{kumar2021recurrent} used Recurrent Neural Networks and reinforcement learning models for prediction. They constructed a model using the Modified Long Short-Term Memory to predict new infections, deaths, and recoveries in the near future. Deep learning reinforcement was proposed to optimize COVID-19 predictions based on symptoms. Real-world data analysis showed promising results, outperforming classicial LSTM and Logistic Regression models in terms of error rate.



Regarding the analysis of COVID-19 using image data, Bouchareb et al. \cite{bouchareb2021artificial} conducted a comprehensive review in 2021, highlighting the use of AI-based models with chest X-ray and computed tomography imaging modalities. The authors emphasized that hybrid methods, which integrate AI-based models with explicit radiometric features, show significant promise for both the diagnosis and prognosis of COVID-19. In a separate study, Ramadhan and Baykara \cite{ramadhan2022novel} proposed an approach to classify patient conditions into three categories: COVID-19, normal, and pneumonia. This method utilized image cropping techniques combined with deep learning, specifically an updated VGG16 convolutional neural network model, applied to X-ray samples.


After recovering from COVID-19, some patients were still affected by long-term symptoms which influenced their physical and cognitive functions. This syndrome has been named ``Long COVID". At the moment, the causes and the duration of long COVID remains not very clear. It is worth mentioning that later in this paper, we will review the notion of long COVID based on WHO definition.

There are a limited number of studies that have explored machine learning's potential in predicting long COVID \cite{sudre2021symptom, kessler2023predictive}. A study by Sudre et al. \cite{sudre2021symptom} used data from over 4,000 individuals to develop an unsupervised times series clustering model predicting long COVID likelihood based on demographic, clinical, and lab features such as age, sex, Body Mass Index (BMI), initial symptoms, disease severity, inflammatory markers like C-reactive protein, and cardiac biomarkers like troponin. Their models also identified specific symptoms and clinical features associated with long COVID, identifying six distinct symptom presentation clusters.

One major challenge in studying this subject is the lack of data. As an evolving crisis, COVID-19 datasets had to be created and collected in real time with limited understanding of the virus. Thus, most data were collected retrospectively from incomplete patient medical files or clinical cohorts of hospitalized patients. However, data suggest that most people affected by Long Covid were never hospitalized, and symptom impact can be greater for non-hospitalized groups. There's often limited knowledge of participants' pre-existing conditions, making it hard to verify that persistent symptoms are new and attributable to COVID-19. Cohorts created during the pandemic often focused on recovery and prognosis, lacking control groups of non-infected participants.


This study utilizes a unique dataset collected and maintained by Lifelines that addresses some of these concerns. Lifelines is a multi-disciplinary, prospective cohort study examining the health and health-related behaviors of 167,729 individuals in Northern Netherlands over three generations. It assesses biomedical, socio-demographic, behavioral, physical, and psychological factors, with a focus on multi-morbidity and complex genetics.

From April 2020 to November 2022, a COVID-19 specific branch involved 30 questionnaires sent to Lifelines adult participants without inclusion criteria. Frequency varied from weekly to bi-monthly. 70,000 participants answered at least one questionnaire, while 11,000 answered all 30. The cohort's duration and size provide valuable data on pre-existing conditions, control groups, and factors influencing Long COVID's emergence, evolution, and severity. Notably, the questions include information on symptoms, infections, and vaccinations.
In this study, we aim to perform data analysis using machine learning techniques to address the following critical research question: \textit{“Can specific pre-infection parameters be identified to predict the severity of Long COVID?”}. The ability to predict Long COVID and identify relevant pre-infection symptoms and parameters holds significant societal implications, impacting mental health, physical well-being, daily functioning, and productivity. To facilitate this, we introduced the concept of Post COVID-19 Symptom Intensity (PCSI) as a measure of symptom persistence following COVID-19 infection. Leveraging various machine learning models, we focused on predicting PCSI using demographic and clinical features. The primary contributions of this work are summarized as follows:

\begin{itemize}
    \item Conducting a comprehensive statistical analysis to identify influential factors associated with the study of long COVID;
    \item Performing predictive analysis of \textit{Post COVID-19 Symptom Intensity} using data-driven approaches;
    \item Interpreting and analyzing the impact of diverse variables on \textit{Post COVID-19 Symptom Intensity}, offering valuable information for medical decision-making;
    \item Developing a Python package\footnote{\url{https://github.com/Mleyliabadi/ML4HEALTH}} for evaluating ML algorithms on health-related (Lifelines) datasets, facilitating reproducibility and further research in the domain.
\end{itemize}

The remainder of this article is structured as follows. Section 2 describes the data preprocessing steps and provides statistical insights into the dataset. Section 3 presents the methodology for predicting long COVID intensity, along with the results and an analysis of key influential factors identified by each model. Finally, Section 4 concludes with a discussion of the study's limitations, societal implications, and future research directions.

\section{Preprocessing and data analysis}
This section presents the data utilized for the analysis and describes the preprocessing steps undertaken to format the data suitably. Additionally, it includes a preliminary statistical analysis to reveal global tendencies. 

\subsection{Data description}
The dataset comprises two main types of variables:

\begin{itemize}
\item Static Variables: These denote fixed attributes of individuals, recorded as single entries in the database. Examples include age, gender, COVID-19 variant, income, smoking status, overall health status, presence of chronic diseases, vaccination status, and the time between vaccination and infection.

\item Dynamic Variables: These variables capture the presence and intensity of symptoms at different time intervals (before, during, and after COVID-19 infection). Symptoms include headache, dizziness, heart or chest pain, lower back pain, nausea, muscle pain, difficulty breathing, feeling warm or cold, numbness or tingling, sore throat, dry or wet cough, fever, diarrhea, loss of smell or taste, and sneezing, among others.
\end{itemize}

Several challenges emerged while working with the data. Similar to many questionnaire-based datasets, there were considerable amounts of missing or aberrant data. Additionally, since the data was collected during an active epidemic, the scope and phrasing of the questionnaires evolved over time, resulting in inconsistencies. Extensive preprocessing was undertaken to address these issues, standardizing the dataset and ensuring a uniform structure suitable for analysis.

\subsection{Definition of Post COVID-19 symptoms intensity (PCSI)}

Post COVID-19 syndrome (Long COVID) refers to a systemic condition where individuals continue to experience persistent symptoms after recovering from a COVID-19 infection (caused by SARS-CoV-2). Although the WHO offers a general definition, it does not specify which symptoms or measurement methods should be considered \cite{soriano_clinical_2022, srikanth_identification_2023}. Consequently, variations in reporting emerge based on the time frame, types of symptoms, and symptom intensity used across studies. In this study, we adhered to the WHO time frame definition: persistent symptoms that cannot be explained by an alternative diagnosis, occurring three months after a confirmed or suspected COVID-19 infection, and lasting for at least two months. Our selection of symptoms was based on the 10 main Long COVID symptoms identified in previous research using the same data set \cite{ballering2022persistence}.

\begin{figure}[htbp]
    \centering
    \includegraphics[width=\columnwidth]{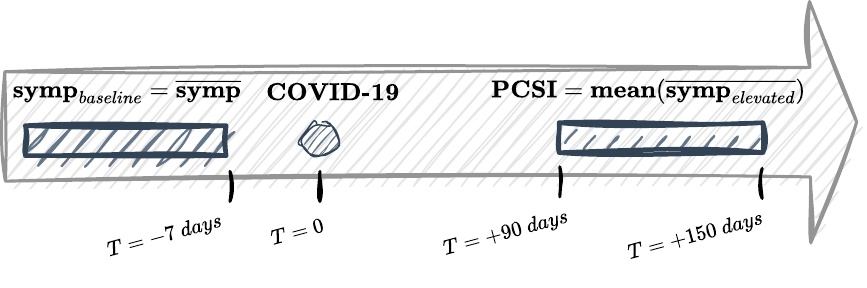}
    \caption{The overall process for defining Post COVID-19 symptom intensity (PCSI) using symptoms (symp) scores. All analyses were centered around the time of the first reported COVID-19 infection.}
    \label{fig:definition_LC}
\end{figure}

Symptom intensity was measured using a 5-point Likert scale, reflecting the degree to which participants were affected (1 = not at all, 5 = extremely) during the seven days preceding the completion of the questionnaire (see Figure \ref{fig:definition_LC}). A symptom was considered present if its score was at least 3 (moderately affected). A baseline profile was established for each participant by calculating the mean symptom intensity across all available questionnaires completed at least seven days before infection. Participants who did not have data for this period were excluded from the analysis.

To determine the persistence of post-COVID symptoms, the mean score for each symptom was calculated for the 90-to-150-day period following infection. A patient was classified as Long COVID positive if they reported at least one persistent symptom (mean score $\geq$ 3) that also increased by at least one point compared to their baseline. 

We defined \textit{Post COVID-19 Symptoms Intensity (PCSI)} as a continuous measure of how participants were impacted by persistent symptoms attributable to COVID-19. PCSI was calculated as the average score of all symptoms that met the Long COVID criteria (impactful symptoms with scores $\geq$ 3, elevated by at least one point compared to baseline). Using a continuous variable, such as PCSI, offers several advantages over a categorical one: it retains more granular information about the variation in symptom severity. This allows for a more nuanced understanding of the relationships between predictors and outcomes. Moreover, this is better suited for statistical and machine learning models that can leverage the continuous nature of the data. In addition, PCSI can also act as a proxy for the categorical Long COVID definition when necessary, providing flexibility for different analytical approaches.

\subsection{Data cleaning and processing}
\begin{figure}[htbp]
    \centering
    \includegraphics[width=\columnwidth]{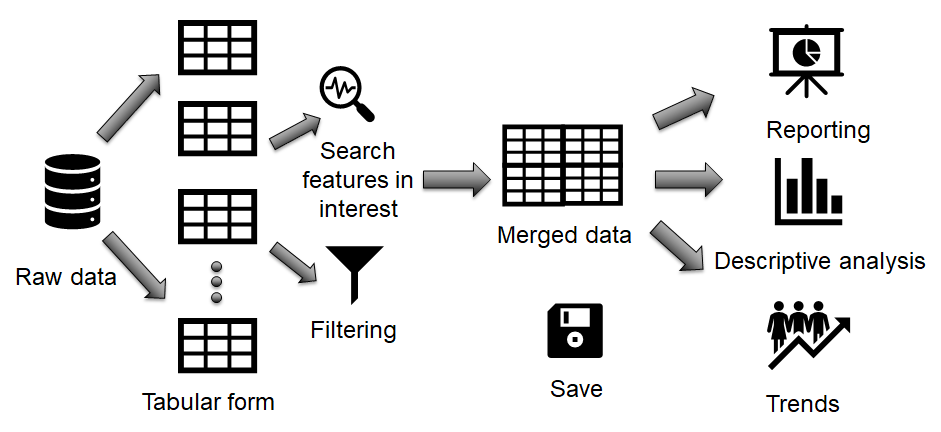}
    \caption{Data pre-processing and extraction strategy}
    \label{fig:scheme_data_preprocessing}
\end{figure}

The diagram in Figure \ref{fig:scheme_data_preprocessing} outlines the methodology employed for processing, extracting, and analyzing the data. As depicted, the raw data were organized into multiple tables, each containing information collected at the participant level for specific dates. After cleaning and preprocessing these tables, participants with a sufficient number of shared variables were filtered. This filtering process resulted in the creation of a merged database that consolidated all the necessary information required for the study and analysis.

\begin{figure}[htbp]
    \centering
    \includegraphics[width=0.9\columnwidth]{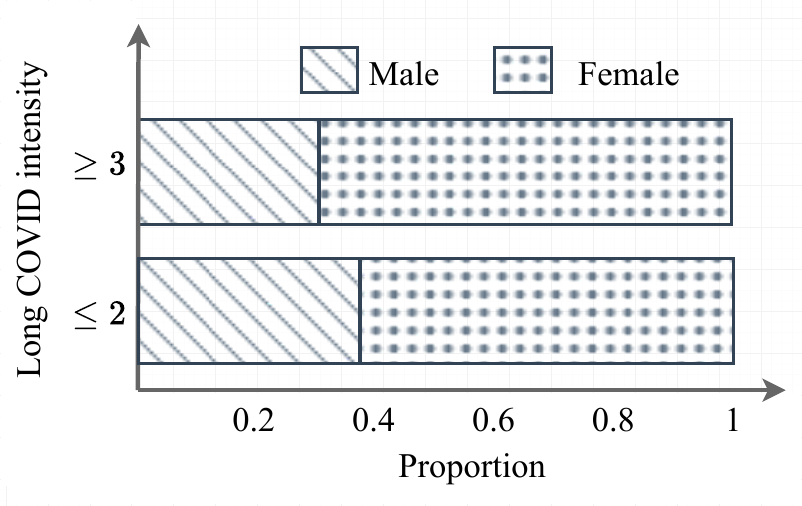}
    \caption{Bar chart demonstrating the gender proportion with respect to low ($\leq 2$) and high ($\geq 3$) intensities of long COVID.}
    \vspace{-0.4cm}
    \label{fig:bar_chart}
\end{figure}

For the predictive analysis, we adopted the steady-state hypothesis, utilizing only the pre-infection period for feature extraction. After preprocessing the dataset to remove incomplete observations for the required factors, a total of 4,657 patients were included in this study. Figure \ref{fig:bar_chart} illustrates the distribution of men and women in the dataset based on low ($\leq2$) and moderate-to-high ($\geq3$) intensities of Post COVID (PC) symptoms, displayed using a bar chart. For the high intensity of PC symptoms, women account for 72\% of the cases, indicating that women are more likely to be at risk for long COVID than men. Conversely, for low PC symptom intensities, the proportion of women is smaller. It is also worth noting that women constitute 64\% of the entire dataset.

\subsection{Preliminary statistics}
To assess the impact of input variables and investigate potential dependencies between the input variables and the outcome (presence of long COVID), we applied two statistical tests. These tests are outlined below:

\begin{itemize}
  \item \textit{Chi-square test}: This test assesses whether two categorical variables are independent \cite{mchugh2013chi}. By evaluating the p-value obtained from the test statistic at the chosen confidence level, we determine whether to reject the null hypothesis in favor of the alternative hypothesis. A confidence level of 95\% is typically used and the null hypothesis is rejected if $p-value<0.05$. 
  \item \textit{Cramer's V test}: This test quantifies the strength of association between two categorical variables \cite{cramer1999mathematical}. A value close to zero indicates a weak dependency, while a value approaching 1 suggests a strong dependency.
\end{itemize}

\begin{figure}[htpb]
    \centering
    \includegraphics[width=0.91\columnwidth]{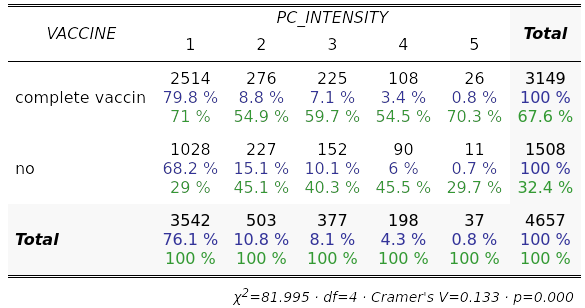}
    \caption{Chi-square test between vaccination and long COVID intensity. The test results indicate a significant relationship ($p < 0.05$) between vaccination and long COVID intensity.}
    \vspace{-0.1cm}
    \label{fig:chi_sq_vaccin_COVID}
\end{figure}

Using these tests, we analyzed the influence of vaccination on PC symptom intensity, with the results depicted in Figure \ref{fig:chi_sq_vaccin_COVID}. This analysis was also conducted for other variables; however, we present only the results for vaccination, as it serves as a crucial preventive measure against COVID-19. To simplify the interpretation, we rounded the long COVID intensity. From the figure, it is evident that most patients who are fully vaccinated are less likely to experience high levels of long COVID intensity (2,790 out of 3,149 or 88\% vaccinated patients report intensity levels 1 or 2). However, due to a lack of representative observations for higher intensity levels, we cannot confidently establish a relationship between vaccination and long COVID intensity for these cases. The Chi-square test statistic ($p < 0.05$) confirms the significance of this relationship, even though the strength of the association is weak (Cramer's $V = 0.072$). 

\begin{figure}[htbp]
     \centering
     \includegraphics[width=\columnwidth]{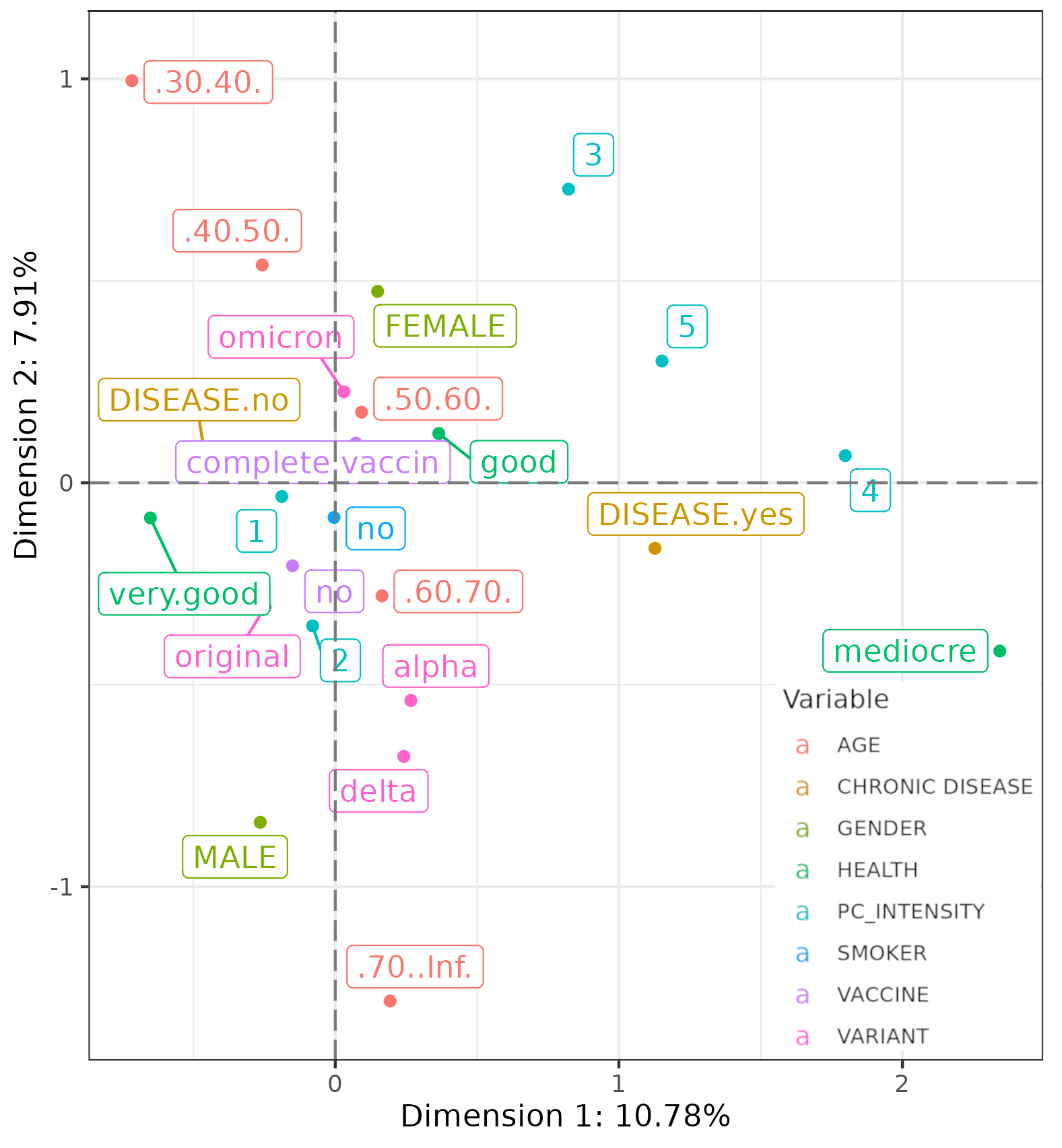}
    \caption{Multiple Correspondence Analysis applied to static and vaccination variables. Each axis represents a component, with the corresponding variance explained. The Post-COVID (PC) intensity variable is discretized by rounding continuous values to the nearest integer (1, 2, 3, 4, and 5 modalities, shown in clear blue).}
\label{fig:MCA_test}
\end{figure}

The final step in our descriptive analysis involved examining the relationships between multiple variables simultaneously using Multiple Correspondence Analysis (MCA) \cite{greenacre2006multiple}. MCA is a method used to identify and visualize underlying structures in a set of nominal categorical data. It can be seen as the categorical equivalent of principal component analysis, projecting data points into a low-dimensional Euclidean space \cite{greenacre2006multiple}. The MCA results are shown in Figure \ref{fig:MCA_test}.

The MCA plot reveals that high long COVID intensity (5) is linked to the presence of chronic diseases and poorer overall health. Additionally, it appears that women are more likely to experience higher long COVID intensity compared to men. The original COVID variant does not show a strong correlation with long COVID, suggesting a lower risk. Lastly, individuals in better general health seem to have a reduced risk of developing long COVID.

\section{Methodology and results}

\subsection{Predictive Analysis and Benchmark Methodology}

\begin{figure}[h]
    \centering
    \includegraphics[width=\columnwidth]{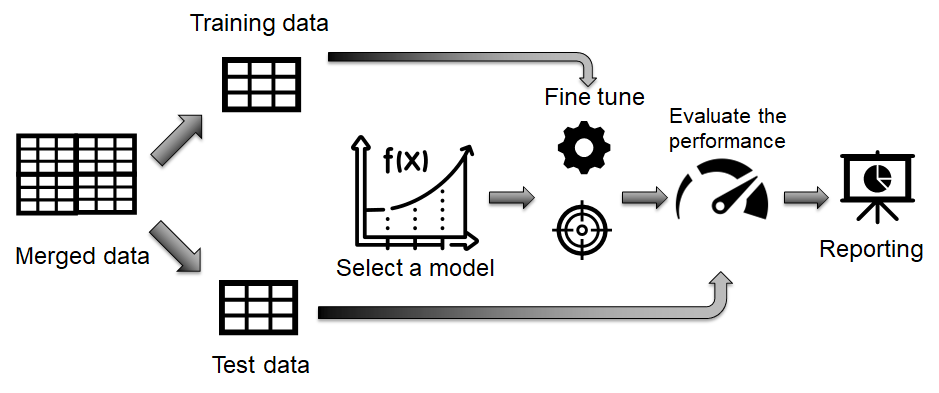}
    \caption{Benchmark and evaluation pipeline}
    \label{fig:scheme_method}
\end{figure}

In this section, we outline an evaluation pipeline designed to select and benchmark various predictive models using the data obtained from the pre-processing stage. The goal of this study is to predict the target variable, \( y \), which represents the intensity of long COVID. The intensity is modeled as a continuous variable ranging between 1 (low intensity) and 5 (high intensity). Given its continuous nature, the problem is formulated as a regression task, where the models aim to approximate the mapping \( f: \mathbf{X} \rightarrow y \), with \( \mathbf{X} \in \mathbb{R}^p \) being the set of \( p \) explanatory variables (features). The overall structure of the proposed pipeline is illustrated in Figure \ref{fig:scheme_method}. In the context of statistical learning, the data are partitioned into two subsets:  
\begin{itemize}
    \item \textbf{Training set (\( \mathcal{D}_{\text{train}} \))}: It involves 70\% of all the patients (4657)  are considered as the training set and used to estimate the parameters \( \theta \) of the predictive model \( f_\theta \);
    \item \textbf{Validation set (\( \mathcal{D}_{\text{val}} \))}: It involves 10\% of the training set and used to estimate the hyperparameters \( \theta_{hyp} \) of the predictive model \( f_\theta \);
    \item \textbf{Test set (\( \mathcal{D}_{\text{test}} \))}: It involves 30\% of all the patients, and it is used to evaluate the performance of the trained model on unseen data and assess the generalization ability of the model.  
\end{itemize}  

After selecting the models, their hyperparameters (\( \theta_{\text{hyp}} \)) are fine-tuned to optimize performance. This crucial step enhances the model's predictive capabilities and is elaborated on in Section \ref{Sec:experimental_setup}. The optimization process may involve techniques such as grid search or gradient-free optimization methods (e.g., Nevergrad), depending on the model's complexity.

Subsequently, each model's performance is evaluated based on a set of criteria measuring accuracy and reliability. The results are presented using both tabular and graphical tools to facilitate comparison and interpretation. These results offer insights into the models' predictive capabilities and help identify the most suitable approach for modeling long COVID intensity.

Lastly, to identify patient profiles and implement preventive measures against long COVID syndrome, it is crucial to assess the significance of the explanatory variables used for model training and parameter adjustment. Depending on the model utilized, we employ explanation and interpretation tools to extract meaningful insights. These insights can offer valuable guidance for the medical field.

We also provide a Python package that encompasses the functionalities described in the aforementioned benchmarking steps. Detailed instructions on how to use this framework can be found in our GitHub repository \footnote{\url{https://github.com/Mleyliabadi/ML4HEALTH}}.

\subsection{Evaluated Methods}
\label{Sec: evaluated_methods}

To tackle the regression problem, we evaluated and compared several data-driven models, including Linear Ridge Regression (LR), Random Forest (RF), Gradient Boosting (GB), and Multi-Layer Perceptron (MLP). LR is a linear model enhanced with regularization to address multicollinearity and reduce overfitting. RF is an ensemble technique that builds multiple decision trees and aggregates their predictions for robust regression or classification \cite{ho1995random}. GB sequentially combines weak learners, typically decision trees, to minimize errors and improve predictive accuracy \cite{hastie2009elements}. Finally, MLP, a feed-forward neural network, excels at modeling non-linear relationships with fully connected layers of neurons and non-linear activation functions \cite{cybenko1989approximation}.

\subsection{Evaluation criteria}
\label{Sec:evaluation_criteria}
Considering that long COVID intensity is a continuous target variable, we have chosen the following evaluation criteria to assess the model's performance:  
\begin{itemize}
\item\textit{MAPE (Mean Absolute Percentage Error)} measures forecasting accuracy by calculating the average absolute percentage error between predicted and actual values, providing interpretability but being sensitive to small actual values  
\item\textit{MAE (Mean Absolute Error)} provides a straightforward measure of accuracy by averaging the absolute differences between predictions and actual values, expressed in the same units as the data.  
\item\textit{MSE (Mean Squared Error)} is an average of the squared differences between predictions and actual values, penalizing larger errors more heavily and making it sensitive to outliers.  
\item\textit{Pearson correlation} quantifies the linear relationship between predicted and actual values, ranging from -1 (negative correlation) to +1 (positive correlation), with 0 indicating no linear relationship.
\end{itemize}

\subsection{Experimental setup}
\label{Sec:experimental_setup}
We fine-tuned all the presented models to determine the optimal set of hyperparameters. For hyperparameter optimization, we employed the Nevergrad library \cite{nevergrad}. The best hyperparameters for MLP were: 3 hidden layers with 126 neurons each, ReLU activation function, Adam optimizer with a learning rate of $9\times10^{-4}$, and 200 training epochs. For RF, the optimal settings included 500 estimators, a maximum depth of 12, a maximum sample fraction of 0.4, and 25 maximum features. Similar hyperparameters were achieved for GB. Lastly, for LR, the L2 regularization strength multiplier was set to 1.0.

\begin{figure}[htbp]
    \centering
    \includegraphics[width=0.9\columnwidth]{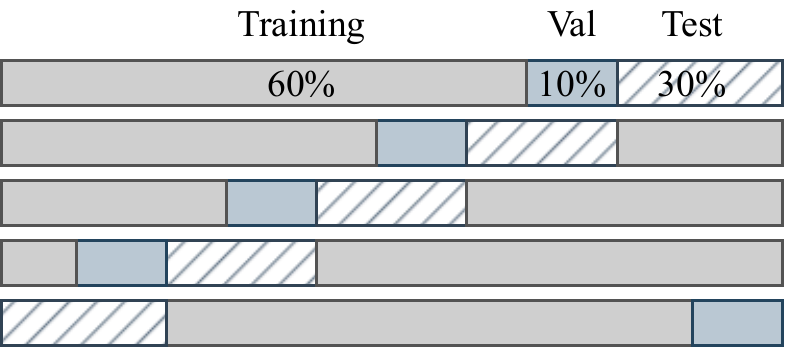}
    \caption{K-fold cross validation ($K=5$) strategy adopted for training and evaluation of models}
    \vspace{-0.4cm}
    \label{fig:cross_val}
\end{figure}

To ensure reliable performance, we utilized \( K \)-fold cross-validation on the training set (\( K = 5 \) in this study). This process splits \( \mathcal{D}_{\text{train}} \) into \( K \) equal parts, sequentially using \( K-1 \) folds for training and one fold for validation (illustrated in Figure \ref{fig:cross_val}). The average performance across folds provides a robust estimate of model accuracy and stability. We ensured that the train and test sets maintained a balanced distribution of patients across various intensity levels in all cross-validation partitions.

\subsection{Results}
In this section, we present and analyze the results obtained by various data-driven approaches for predicting long COVID intensity. The outcomes are summarized in Table \ref{tbl:comparison_regression_complete}, where different combinations of characteristics are evaluated using the four aforementioned methods (see Section \ref{Sec: evaluated_methods}) and assessed through multiple evaluation criteria (see Section \ref{Sec:evaluation_criteria}). The ``All" feature combination represents the integration of all characteristics, including static variables, symptoms, and vaccination data. For clarity, the best results for each method are marked in bold, while the best performance for each evaluation criterion is highlighted in green. Additionally, all performance metrics are averaged across \( K = 5 \)-fold cross-validation (refer to Section \ref{Sec:experimental_setup} for details on the experimental setups).

\begin{table}[hbtp]
\centering
\caption{Comparison between various introduced models and features combination for prediction of \textit{long COVID intensity}. Various metrics are used for evaluation and results are reported as (mean $\pm$ standard deviation) through 5-fold cross-validation. Pearson correlation is reported using the pair (test statistic, p-value). \deleted[id=ML]{The best performance for each method is shown in bold and the best performances obtained through all the methods are highlighted using green color.}}
\resizebox{\columnwidth}{!}{
\begin{tabular}{ll|llll}
\toprule\hline
\multicolumn{2}{l|}{\cellcolor[HTML]{C0C0C0}} & \multicolumn{4}{c}{Evaluation criteria} \\ \hline
\multicolumn{1}{c|}{Methods} & Features & \multicolumn{1}{c|}{MAE} & \multicolumn{1}{c|}{MSE} & \multicolumn{1}{c|}{MAPE} & \multicolumn{1}{c}{Pearson} \\ \hline
\multicolumn{1}{c|}{} & All &  \multicolumn{1}{c|}{\textbf{.61 $\pm$ .01}} & \multicolumn{1}{c|}{\textbf{.68 $\pm$ .02}} & \multicolumn{1}{c|}{\textbf{.29 $\pm$ .01}} & \multicolumn{1}{c}{\textbf{(.56, 6e-70)}}\\
\multicolumn{1}{c|}{} & Static &  \multicolumn{1}{c|}{.71 $\pm$ .02} & \multicolumn{1}{c|}{.91 $\pm$.05} & \multicolumn{1}{c|}{.35 $\pm$ .01} & \multicolumn{1}{c}{(.28, 2e-16)}\\
\multicolumn{1}{c|}{} & Symptoms &  \multicolumn{1}{c|}{.62 $\pm$ .02} & \multicolumn{1}{c|}{.70 $\pm$ .04} & \multicolumn{1}{c|}{.30 $\pm$ .01} & \multicolumn{1}{c}{(.57, 2e-69)} \\
\multicolumn{1}{c|}{\multirow{-4}{*}{LR}} & Vaccination &  \multicolumn{1}{c|}{.81 $\pm$ .02} & \multicolumn{1}{c|}{.99 $\pm$ .05} & \multicolumn{1}{c|}{.41 $\pm$ .01} & \multicolumn{1}{c}{NaN} \\ \hline
\multicolumn{1}{c|}{} & All &  \multicolumn{1}{c|}{\textbf{.60 $\pm$ .01}} & \multicolumn{1}{c|}{\textbf{.67 $\pm$ .02}} & \multicolumn{1}{c|}{\textbf{.28 $\pm$ .01}} & \multicolumn{1}{c}{\cellcolor[HTML]{8DC73E}\textbf{(.58, 7e-73)}}\\
\multicolumn{1}{c|}{} & Static &  \multicolumn{1}{c|}{.72 $\pm$ .02} & \multicolumn{1}{c|}{.93 $\pm$ .05} & \multicolumn{1}{c|}{.35 $\pm$ .01} & \multicolumn{1}{c}{(.26, 1e-15)}\\
\multicolumn{1}{c|}{} & Symptoms &  \multicolumn{1}{c|}{\textbf{.60 $\pm$ .01}} & \multicolumn{1}{c|}{.66 $\pm$ .03} & \multicolumn{1}{c|}{\textbf{.28 $\pm$ .01}} & \multicolumn{1}{c}{(.57, 5e-72)}\\
\multicolumn{1}{c|}{\multirow{-4}{*}{RF}} & Vaccination &  \multicolumn{1}{c|}{.79 $\pm$ .02} & \multicolumn{1}{c|}{.99 $\pm$ .06} & \multicolumn{1}{c|}{.39 $\pm$ .01} & \multicolumn{1}{c}{(.04, 1e-1))}\\ \hline
\multicolumn{1}{c|}{} & All &  \multicolumn{1}{c|}{\textbf{.61 $\pm$ .01}} & \multicolumn{1}{c|}{\cellcolor[HTML]{8DC73E}\textbf{.66 $\pm$ .01}} & \multicolumn{1}{c|}{\textbf{.28 $\pm$ .01}} & \multicolumn{1}{c}{\textbf{(.57, 4e-74)}}\\
\multicolumn{1}{c|}{} & Static &  \multicolumn{1}{c|}{.72 $\pm$ .02} & \multicolumn{1}{c|}{.90 $\pm$ .05} & \multicolumn{1}{c|}{.35 $\pm$ .01} & \multicolumn{1}{c}{(.29, 7e-17)}\\
\multicolumn{1}{c|}{} & Symptoms &  \multicolumn{1}{c|}{\textbf{.61 $\pm$ .01}} & \multicolumn{1}{c|}{.68 $\pm$ .02} & \multicolumn{1}{c|}{.28 $\pm$ .01} & \multicolumn{1}{c}{(.55, 8e-82)}\\
\multicolumn{1}{c|}{\multirow{-4}{*}{GB}} & Vaccination &  \multicolumn{1}{c|}{.81 $\pm$ .02} & \multicolumn{1}{c|}{.99 $\pm$ .06} & \multicolumn{1}{c|}{.41 $\pm$ .01} & \multicolumn{1}{c}{(.05, 6e-1)}\\ \hline
\multicolumn{1}{c|}{} & All &  \multicolumn{1}{c|}{\cellcolor[HTML]{8DC73E}\textbf{.45 $\pm$ .05}} & \multicolumn{1}{c|}{\textbf{.90 $\pm$ .12}} & \multicolumn{1}{c|}{\cellcolor[HTML]{8DC73E}\textbf{.19 $\pm$ .03}} & \multicolumn{1}{c}{\textbf{(.25, 3e-18)}}\\
\multicolumn{1}{c|}{} & Static &  \multicolumn{1}{c|}{.87 $\pm$ .18} & \multicolumn{1}{c|}{1.4 $\pm$ .78} & \multicolumn{1}{c|}{.43 $\pm$ .07} & \multicolumn{1}{c}{(.21, 4e-9)}\\
\multicolumn{1}{c|}{} & Symptoms &  \multicolumn{1}{c|}{.76 $\pm$ .11} & \multicolumn{1}{c|}{.98 $\pm$ .38} & \multicolumn{1}{c|}{.34 $\pm$ .05} & \multicolumn{1}{c}{(.43, 5e-33)}\\
\multicolumn{1}{c|}{\multirow{-4}{*}{MLP}} & Vaccination &  \multicolumn{1}{c|}{.80 $\pm$ .03} & \multicolumn{1}{c|}{1.03 $\pm$ .05} & \multicolumn{1}{c|}{.41 $\pm$ .03} & \multicolumn{1}{c}{(.04, 2e-1)}\\
\bottomrule
\end{tabular}
}
\label{tbl:comparison_regression_complete}
\end{table}

As shown in Table \ref{tbl:comparison_regression_complete}, the best performance for each method is achieved when all features are combined. However, with the exception of MLP, the performance remains comparable even when only symptom-based features are used. It is worth noting that neural network-based methods, such as MLP, have the capability for automatic feature extraction, whereas traditional statistical approaches like LR, RF, and GB require a dedicated feature engineering step.

We observe that the performance, in terms of the MAE metric, remains very similar across the four approaches when all features are combined. An MAE value of \( 0.60 \) indicates that, on average, the predicted values deviate by \( 0.60 \) points from the actual observations. 
Given that the long COVID intensity ranges from 1 to 5, a deviation of \( 0.60 \) in intensity is unlikely to significantly affect the overall conclusions.

Finally, we note that the best result in terms of MAPE is achieved using MLP, with a value of \( 0.26 \). This indicates that, on average, the predictions deviate by 26\% from the actual intensity values. Interestingly, the highest Pearson correlations between predictions and actual values are obtained with RF and GB, rather than MLP. This discrepancy can be attributed to the differences in how these models capture relationships within the data. RF and GB are ensemble-based methods that excel in capturing complex interactions between features, which may result in higher linear correlations (as measured by Pearson correlation) between predicted and actual values. On the other hand, MLP, being a neural network, is better suited for non-linear patterns and optimization for specific loss functions, which may explain its superior performance in minimizing relative errors (as captured by MAPE).

The evaluation criteria presented here allow us to effectively differentiate between the methods based on their predictive performance. However, in the medical field, it is often equally important to understand the impact of explanatory variables on the predictions. In the following section, we delve into model interpretation to gain insights into the factors influencing the predictions and to provide meaningful explanations for the observed outcomes.

\subsection{Interpretation}
\label{Sec: Interpretation}
In this section, the trained models from the previous section are interpreted and explained using explainability tools and the results are presented using numerical tables, visual tools and graphics.

Table \ref{tbl:lr_coef} presents the top 9 most influential features, along with their corresponding Linear Ridge Regression (LR) coefficients, averaged over 5-fold cross-validation. These coefficients indicate the direction and magnitude of each feature's contribution to the prediction of long COVID intensity. Many common COVID symptoms, such as loss of sense of smell, headache, and muscle pain, exhibit strong positive contributions, suggesting they are associated with a higher risk of long COVID. Conversely, certain acute symptoms like fever or pain when breathing show significant negative contributions, indicating that their presence is less likely to increase the risk of long COVID. This distinction highlights the nuanced relationship between acute and long-term COVID symptoms in influencing long COVID outcomes.

\begin{table}[htbp]
    \centering
    \caption{Estimated Coefficients of linear regression for prediction of long COVID}
    \resizebox{\columnwidth}{!}{
    \begin{tabular}{cc|cc}
         \toprule
         \textbf{Variable} & \textbf{Coef} & \textbf{Variable} & \textbf{Coef} \\ \hline\hline
         Loss of sense of smell/taste & 0.32 & Pain when breathing & -0.58\\
         Headache & 0.28 & Fever ($38^{\circ}$  or higher) & -0.27\\
         Muscle pain/aches & 0.27 & Omicron variant & -0.26\\
         Lower back pain & 0.23 & Heaviness in arms/legs & -0.08\\
         Original variant & 0.17 & Very good health & -0.07\\
         Feeling warm \& cold & 0.16 & No chronic disease & -0.07\\
         Red, painful eyes & 0.16 & AGE & -0.06\\
         Sneezing & 0.16 & Smoker & -0.05\\
         Difficulty breathing & 0.14 & Male & -0.03\\
         \bottomrule
    \end{tabular}}
    \vspace{-0.3cm}
    \label{tbl:lr_coef}
\end{table}

\begin{figure}[hbtp]
    \centering
    \includegraphics[width=\columnwidth]{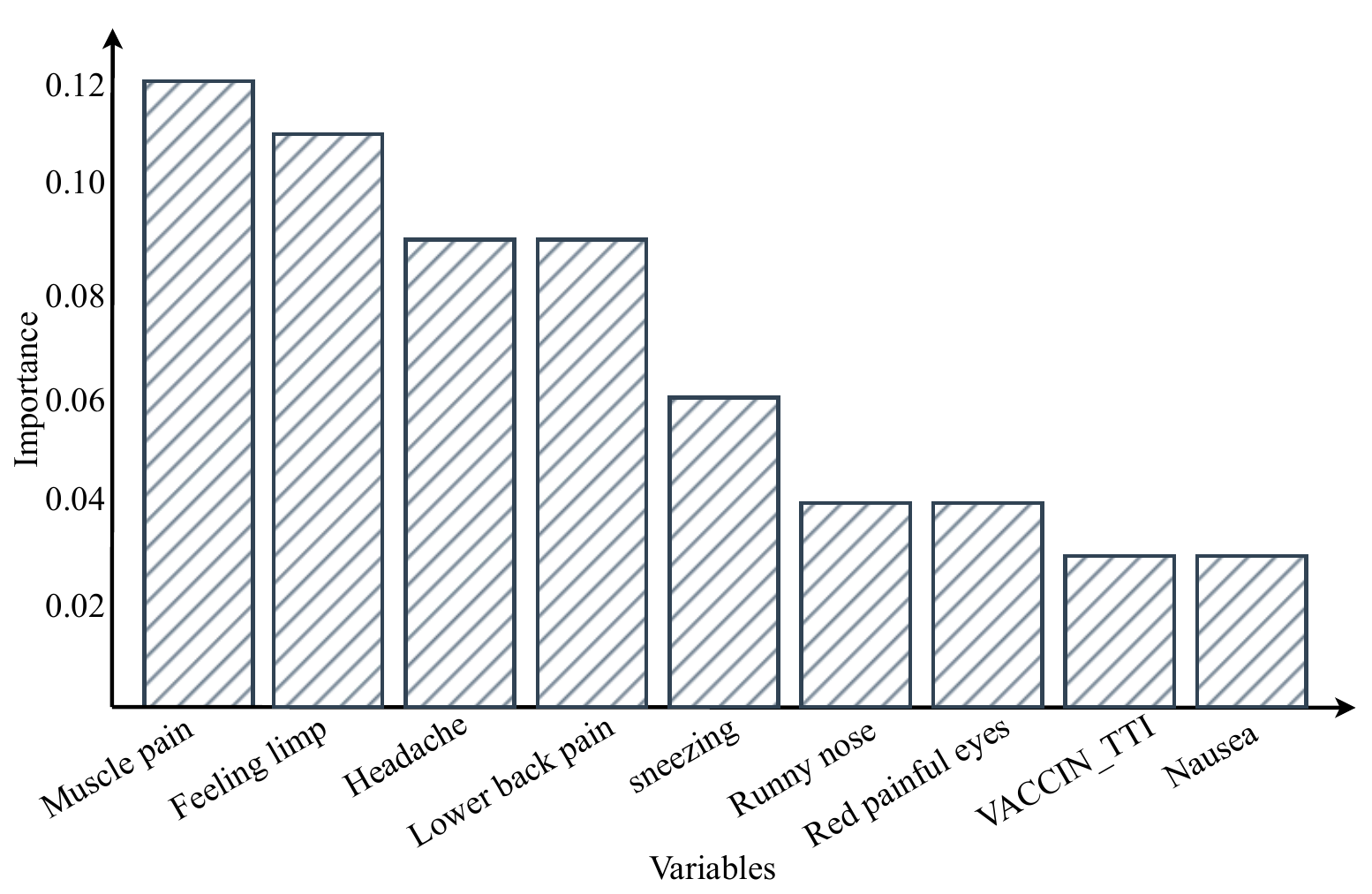}
    \caption{Feature importances resulted using Random Forest model for prediction of long COVID intensity}
    \vspace{-0.2cm}
    \label{fig:important_features}
\end{figure}

Using the Random Forest (RF) model, we illustrated the feature importances with a bar plot in Figure \ref{fig:important_features}. For clarity and brevity, only the top 10 most important features were extracted from the full set. The identified features show some overlap with those presented in Table \ref{tbl:lr_coef}, although their relative importance differs. Notably, muscle pain emerges as the most important predictor of long COVID intensity. Additionally, the feature representing the time interval between vaccination and infection (VACCIN\_TTI in the bar plot) is highlighted as a significant contributor. This finding supports the hypothesis that vaccination timing influences the risk and severity of long COVID, emphasizing its potential impact on disease outcomes.
\vspace{-0.5cm}

\begin{figure}[hbtp]
    \centering
    \includegraphics[width=0.9\columnwidth]{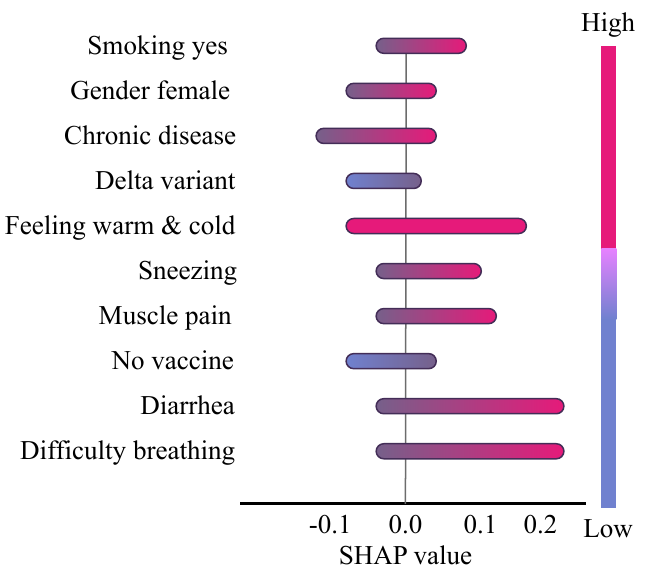}
    \caption{Interpreting MLP influential factors using SHAP explainability tool and extracting the most influential factors}
    \vspace{-0.3cm}
    \label{fig:shap_mlp}
\end{figure}

Utilizing the SHAP explanation tool, we analyzed the influential factors for the MLP model, with the results visualized in Figure \ref{fig:shap_mlp}. The variables with the highest SHAP values were identified as the most influential features in predicting long COVID intensity. Among the top contributors, symptoms such as difficulty breathing, diarrhea, feeling alternately warm and cold, muscle pain, and sneezing exhibit positive SHAP values, indicating their significant role in increasing long COVID intensity. This aligns with clinical observations, as these symptoms are commonly linked to both acute and lingering effects of COVID-19. 

Smoking has a positive influence on long COVID intensity, indicating that smokers may experience more severe post-COVID symptoms, likely due to compromised respiratory health. Conversely, the absence of chronic diseases is associated with reduced long COVID intensity, reflecting the importance of overall baseline health in mitigating long-term effects. Vaccination is another key factor, with lower values of the "no vaccine" variable correlating with reduced long COVID intensity. This supports the growing body of evidence that vaccination not only protects against severe acute COVID-19 but also reduces the risk of persistent symptoms. Lastly, the analysis reveals that females may be more susceptible to long COVID, which is consistent with findings from other studies suggesting a higher prevalence of post-viral syndromes in women, potentially linked to hormonal or immunological differences\cite{bai2022female}. This comprehensive insight underscores the multifaceted nature of long COVID and highlights the importance of individual factors in shaping its intensity.

\section{Discussion and Conclusion}
This study aimed to identify patient profiles at higher risk of developing long COVID and predict its intensity using machine learning approaches. We utilized features that were grouped into static, vaccination, and symptom-related variables. Statistical analyses revealed that women and patients with chronic diseases are more susceptible to long COVID. Predictive analysis using four different models demonstrated strong performance across all methods when combining all features, with MLP showing slightly better results in terms of MAPE. The interpretability analyses identified key predictors, including loss of smell, headache, muscle pain, and vaccination timing, as well as protective factors like the absence of chronic diseases. These insights provide valuable information for tailoring interventions and understanding the underlying risk factors of long COVID.

\par\smallskip
\noindent\textbf{Limitations.}  
The steady-state assumption in our analysis limits the ability to capture temporal relationships between symptoms or events. Model performance is also constrained by the quality and completeness of the dataset, highlighting the need for validation on independent datasets to ensure robustness in real-world scenarios. Additionally, while the models offer predictive value, they are intended as tools to complement clinical judgment rather than replace it.

\par\smallskip
\noindent\textbf{Societal Impact.}  
Long COVID has profound societal implications, affecting mental health, physical well-being, daily functioning, and productivity \cite{mayhew2021coronavirus}, \cite{maclean2023impact}. It disrupts educational and professional activities, with children and adults experiencing isolation, stress, and cognitive impairments. Predicting long COVID intensity can inform early interventions, alleviate healthcare burdens, and improve patients' quality of life.

\par\smallskip
\noindent\textbf{Future Work.}  
To address the dataset's challenges, future research will focus on improving data quality and incorporating temporal dynamics, such as symptom progression and vaccination timelines, using time-series models like RNNs. Additionally, introducing uncertainty quantification will enhance prediction reliability and enable more actionable and trustworthy decision-making in clinical settings.

\section*{Acknowledgment}
The Lifelines Corona Research Initiative collected the data. Special thanks to our UMCG partners and ZonMw COVID-19 Program who also supported this work.

\printbibliography

\end{document}